\documentclass[10pt,twocolumn,letterpaper]{article}

\usepackage{iccv}              
\usepackage{multicol}
\usepackage{multirow}
\usepackage{pifont}
\usepackage{colortbl}
\usepackage{graphicx}
\usepackage{amsmath}
\usepackage{algorithm}
\usepackage{algorithmic}
\usepackage{graphicx}
\newcommand*{\affaddr}[1]{#1} 
\newcommand*{\affmark}[1][*]{\textsuperscript{#1}}
\newcommand*{\email}[1]{\texttt{#1}}

\definecolor{iccvblue}{rgb}{0.21,0.49,0.74}
\usepackage[pagebackref,breaklinks,colorlinks,allcolors=iccvblue]{hyperref}

\def\paperID{15120} 
\def\confName{ICCV}
\def\confYear{2025}

\title{V\textsuperscript{2}Flow: Unifying Visual Tokenization and Large Language Model Vocabularies for Autoregressive Image Generation}

\author{%
{Guiwei Zhang\affmark[1]}, Tianyu Zhang\affmark[1], Mohan Zhou\affmark[1], Yalong Bai\affmark[1]\thanks{Project lead}, Biye Li\affmark[1]\\
\affaddr{\affmark[1]{ Du Xiaoman Financial.}}\\
\small
\email{{ \{zhangguiwei0610, tianyu1949, libiye\}@gmail.com}
\email{\{mhzhou99, ylbai\}@outlook.com}}}

\begin{document}
\maketitle
\begin{abstract}
We propose V\textsuperscript{2}Flow, a novel tokenizer that produces  discrete visual tokens capable of high-fidelity reconstruction,  while ensuring structural  and latent distribution alignment with the vocabulary space of  large language models (LLMs). Leveraging this tight visual-vocabulary coupling, V\textsuperscript{2}Flow enables  autoregresive visual generation on top of   existing  LLMs. Our  approach formulates visual tokenization as a flow-matching problem, aiming to learn a mapping from a standard normal prior to the continuous image distribution, conditioned on token sequences embedded within the  LLM’s vocabulary space.
The effectiveness of  V\textsuperscript{2}Flow  stems from two core designs. First, we propose a \underline{\textbf{V}}isual \underline{\textbf{V}}ocabulary  resampler, which compresses visual data into compact token sequences, with each represented as a soft categorical distribution over LLM's vocabulary. This allows seamless  integration of visual tokens into existing LLMs for autoregressive visual generation. Second, we present a masked autoregressive Rectified-\underline{\textbf{Flow}} decoder, employing a masked transformer encoder-decoder to refine visual tokens into  contextually enriched embeddings. These embeddings then condition a  dedicated velocity field  for precise reconstruction. Additionally, an autoregressive  rectified-flow sampling strategy  is incorporated, ensuring flexible sequence lengths while  preserving competitive reconstruction quality. Extensive experiments show that  V\textsuperscript{2}Flow outperforms mainstream VQ-based tokenizers and facilitates autoregressive visual generation on top of  existing. \footnote{\url{https://github.com/zhangguiwei610/V2Flow}}
\end{abstract}    
\section{Introduction}
\label{sec:intro}

\begin{figure*}[t]
  \centering
   \includegraphics[width=0.95\textwidth]{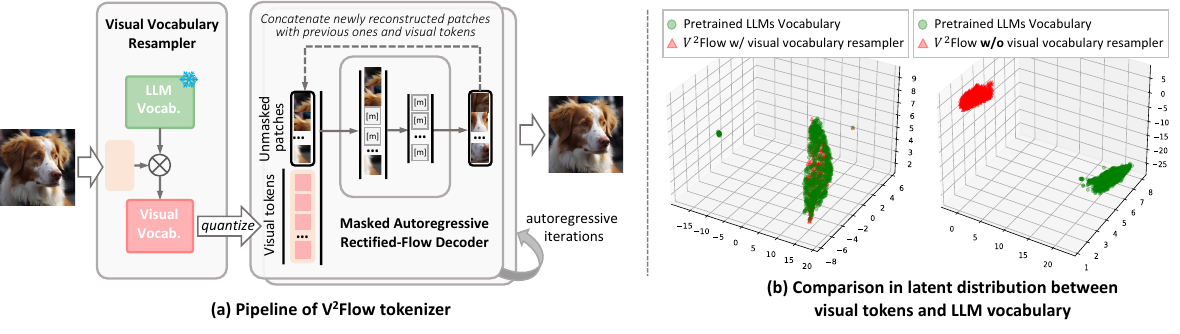}
   \caption{ Highlights of  V\textsuperscript{2}Flow tokenizer. In (a),  a visual vocabulary resampler compresses visual content into a compact one-dimensional token sequence. Each token is directly expressed within the latent distribution of existing LLMs vocabulary space, as illustrated in (b). This design facilitates autoregressive visual generation on top of existing LLMs. Subsequently, the quantized visual tokens condition on a masked autoregressive Rectified-Flow decoder for high-fidelity visual reconstruction. 
   }
\label{fig_intro}
\end{figure*}

Autoregressive modeling has recently emerged as a prominent paradigm  in both   natural language processing~\cite{touvron2023llama,bai2023qwen,dubey2024llama} and multimodal understanding ~\cite{liu2024visual,zhu2023minigpt,zhang2023video,yao2024minicpm,cheng2024videollama},   advancing the integration 
 of visual perception with linguistic interfaces.
Building upon these successes, an increasing body of work~\cite{fang2024puma,sun2023emu,dong2023dreamllm,ge2024seed,sun2024generative} has begun to explore autoregressive modeling  for visual generation.
Although promising, these approaches ~\cite{ge2024seed,sun2024generative,zhou2024transfusion,xie2024show,xiao2024omnigen} operate under a paradigm that representing language in a discrete, categorical form~\cite{touvron2023llama}, while modeling  visual information within continuous feature spaces~\cite{kingma2013auto,rezende2014stochastic}, i.e.,  equipping LLMs with the power of diffusion models~\cite{ho2020denoising,song2020denoising}. This representational discrepancy introduces additional  architectural complexity and computational overhead, preventing autoregressive visual generation from attaining the streamlined efficiency seen in language tasks.

Recent research~\cite{team2024chameleon,sun2024autoregressive,kondratyuk2023videopoet,yu2023language}  has explored one promising solution: leveraging a single transformer architecture to unify visual and textual data within the next-token prediction framework. In this paradigm, 
visual content is quantized  into discrete tokens~\cite{esser2021taming,lee2022autoregressive,zheng2022movq,yu2023magvit} that can be jointly processed alongside categorical, discrete textual data for autoregressive visual generation.
However, a fundamental challenge arises. Existing methods predominantly employ VQ-based  tokenizers~\cite{esser2021taming,shi2024taming} optimized solely for visual reconstruction, causing the distribution  of discretized visual tokens to misalign with the semantically rich representations of text. Furthermore, existing approaches preserve the intrisic two-dimensional structure of images, forcing autogressive models to predict  visual tokens in a line-by-line manner. This  diverges  sharply from the processing of  one-dimensional textual data. Both structural and latent distribution disparities introduce competition between two modalities, thereby hindering the efficiency and effectiveness of unified  autoregressive modeling ~\cite{team2024chameleon}.

These limitations reveal a critical gap within current autoregressive visual generation paradigms: the absence of a visual tokenizer capable of achieving both high-fidelity image reconstruction and effective latent distribution alignment with pretrained LLMs textual features. Addressing this gap is critical for effective multimodal integration and enhanced image-text association learning. Motivated by this, we pose a central question: {\it\textbf{Can we design a visual tokenizer capable of generating discrete tokens that achieve accurate reconstruction and seamless integration with pretrained LLMs' vocabularies?}} 

In response, we propose  V\textsuperscript{2}Flow, a novel tokenizer designed  to generate  discrete  tokens aligned structurally and in latent distribution with existing LLMs. Our key insight is to embed visual features directly into the vocabulary space of pretrained  LLMs (e.g., the \texttt{LLaMA series}~\cite{dubey2024llama,touvron2023llama}). This paradigm naturally situates visual representations within the exisitng text embedding space of LLMs, circumventing the need for forced semantic alignment.  This effectively harmonizes visual and textual modalities, significantly alleviating challenges induced by discrepancies in their respective feature scales and distributions. Consequently, our approach facilitates efficient autoregressive image generation directly leveraging pretrained LLMs.

Although our V\textsuperscript{2}Flow  operates with a quantized vocabulary of limited cardinality, we empirically demonstrate that this constraint does not restrict the representational capability. Instead, the expressive power of V\textsuperscript{2}Flow  primarily emerges from the intricate interactions and rich co-occurrence patterns among visual tokens. Fine-grained visual details, particularly high-frequency structures,  naturally arise through explicit modeling of  token dependencies~\cite{he2022masked,wei2022masked}.  
Inspired by  masked image modeling~\cite{he2022masked,li2024autoregressive}, we introduced a targeted masked approach, designed to  capture these critical co-occurrence patterns. Furthermore, 
we integrate this with Rectified-Flow sampling~\cite{esser2024scaling}, effectively transforming discrete  tokens into continuous distributions for  high-quality visual reconstruction.


\cref{fig_intro} highlights two primary designs of the V\textsuperscript{2}Flow tokenizer. \textbf{First}, to enforce structural alignment and latent distribution consistency between  visual tokens and pretrained LLMs vocabulary, we propose a \emph{visual vocabulary resampler}. This resampler  compresses visual content into a compact one-dimensional token sequence, mapping each token into a soft categorical distribution over the LLM vocabulary space.  
As illustrated in ~\cref{fig_intro}(b), visual tokens produced without our resampler exhibit latent distributions distinctly divergent  from textual embeddings. In contrast, the  tokens processed by our resampler exhibit latent distribution closely aligned with the LLM vocabulary. Both structural and latent distribution compatibility can significantly reduce complexity when integrating  visual tokens directly into existing LLMs. \textbf{Second}, to refine quantized token sequences for high-fidelity reconstruction, we propose a \emph{masked autoregressive rectified-flow decoder}. Inspired by recent advances in masked generative modeling, our decoder employs a masked transformer encoder-decoder design, enriching visual tokens with contextually rich information. 
 These embeddings  subsequently  condition a dedicated velocity field model to reconstruct continuous visual distributions from a standard normal prior. During rectified-flow sampling, our decoder adopts an ``next set-of-tokens'' prediction strategy. Compared to prior tokenizers~\cite{yu2024image} that 
  solely  rely on masked encoder-decoder, this approach facilitates superior reconstruction quality with shorter token sequences, thereby improving overall compression efficiency. (Please see \cref{sec:experiments.rec} for details). Our contributions include:
\begin{itemize}
    \item[$\bullet$] We propose a visual vocabulary resampler that quantizes images into a compact one-dimensional  sequence represented within the LLM vocabulary space. By leveraging these tokens, our tokenizer  facilitates  seamless autogressive visual generation built upon existing LLMs.
    \item[$\bullet$] We develop a masked autoregressive rectified-flow decoder, equipped  with an autoregressive  sampling strategy. This design facilitates visual tokenization at flexible sequence lengths while preserving robust and competitive reconstruction quality.
  \item[$\bullet$] Extensive experimental results show that our V\textsuperscript{2}Flow tokenizer achieves competitive reconstruction performance compared to mainstream VQ-based tokenizers, and effectively promote autoregressive visual generation.
\end{itemize}

\section{V\textsuperscript{2}Flow tokenizer}
\label{sec:methods}
\begin{figure*}[t]
  \centering
   \includegraphics[width=0.95\textwidth]{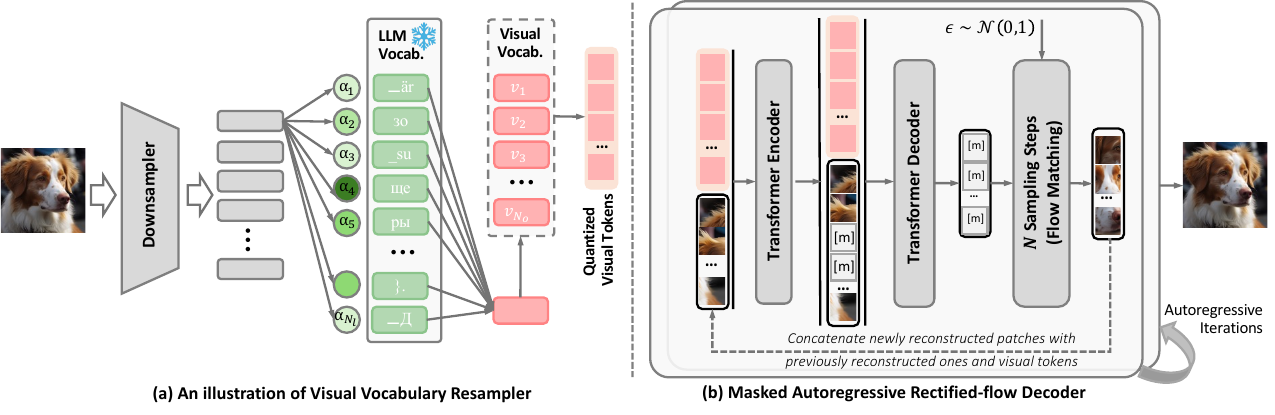}
   \caption{ Overview of the V\textsuperscript{2}Flow tokenizer, including  \ding{202} Visual Vocabulary Resampler and \ding{203} Masked  Autoregressive  Rectified-Flow decoder. The first component is designed to  compress visual content into a compact,  one-dimensional discrete token sequence that are directly expressed within existing LLMs vocabularies. This enables seamless  autoregressive visual generation \textbf{on top of} existing LLMs. Furthermore,  the  Masked Autoregressive Rectified-Flow decoder refines quantized tokens through a masked transformer encoder-decoder, producing visually enriched  embeddings. These embeddings  condition  a  tailored velocity field model  to reconstruct the underlying visual content. Finally, a rectified-flow sampling strategy with autoregressive prediction offers flexibility in sequence length while  preserving competitive reconstruction performance.
   }
   \label{fig:overview_v2flow}
\end{figure*}


\subsection{Task Formulation and Overview}
\textbf{Visual Tokenization from a Flow-Matching Perspective.}
Our goal is to  compress images into compact, one-dimensional sequences of quantized tokens. Each token 
is embeded within the vocabulary space of pretrained LLMs. This tokenization process must satisfy two essential criteria: 

\begin{enumerate}[label=--] \item \emph{High-fidelity visual reconstruction.} The quantized tokens must preserve essential visual details to ensure faithful image reconstruction. \item \emph{Structural and latent distribution alignment with LLMs.} Each visual token should be naturally integrated into LLMs vocabulary space, enabling autoregressive visual generation with minimal adjustments to existing LLMs. \end{enumerate}
To meet these requirements, we formulate visual tokenization as a flow-matching problem. In particular, we learn a mapping from latent variables $\epsilon \sim \mathcal{N}(0, 1)$  to samples $z$ from the  visual data distribution $q$,  via an ordinary differential equation (ODE),
\begin{equation}
\label{ode}
d z_t=\psi_{\Theta}\left(z_t, t,\mathbf{V}^2\right) d t
\end{equation}
where $\psi_{\Theta}$ denotes a learnable velocity field  parameterized by the weights $\Theta$ of our tokenizer decoder, and $t \in[0,1]$ is the time-step. The term $\mathbf{V}^2$ represents the quantized token sequence from our  visual vocabulary resampler (see \cref{sec3.1}), which conditions the velocity field $\psi_{\Theta}$ to guide the generation process. Each token in  $\mathbf{V}^2$ is directly represented within the LLMs vocabulary space. This  effectively bridges the distribution gap between visual tokens and the LLM's generative capability.

Directly solving the ODE in ~\cref{ode} with differentiable ODE solvers is computationally expensive. A more efficient alternative  is to regress a time-dependent vector field $u_t\left(z \mid \epsilon\right)$ that generates a probability path between $\mathcal{N}(0, I)$ and the target distribution $q$.
For optimization efficiency, we employ a rectified-flow~\cite{esser2024scaling}, in which the trajectory between the target distribution and standard normal distribution is assumed to follows a ``straight-line" path:
\begin{equation}
u_t\left(z \mid \epsilon\right) = (1 - t) \cdot z + t \cdot \epsilon
\end{equation}
Overall, the optimization objective of visual tokenization is formulated as minimizing the following flow-matching loss:
\begin{equation}
\mathcal{L}_{FM}=\mathbb{E}_{t, p_t(z \mid \epsilon), p(\epsilon)}\left\|\psi_{\Theta}\left(z_t, t,\mathbf{V}^2\right)-u_t(z \mid \epsilon)\right\|_2^2
\end{equation}
\textbf{Overview of V\textsuperscript{2}Flow.}
We begin by  introducing the proposed visual vocabulary resampler (\cref{sec3.1}), which compresses visual content into a compact token sequence, with each token embedded in the vocabulary space  of existing LLMs.
 Next, in \cref{flow}, we present our masked autoregressive flow-matching decoder, designed to ensure high-quality visual reconstruction. 
 \cref{fig:overview_v2flow} illustrates the overview of the V\textsuperscript{2}Flow tokenizer.
\subsection{Visual Vocabulary Resampler}
\label{sec3.1}
 Building upon recent advances~\cite{cha2024honeybee,cheng2024videollama}, we incorporate  RegStage blocks~\cite{radosavovic2020designing} alongside  a  spatial downsampler  ~\cite{rombach2022high} to compress the input image $X \in \mathbb{R}^{H \times W \times C}$ into  a latent representation $Z\in \mathbb{R}^{h \times w \times d_v}$. Here, $h=H/q,w=W/q$, $q$ is the spatial downsampling factor and $d_v$ denotes the embedding dimension. We then flatten $Z$  into an one-dimensional  sequence $Z_{\operatorname{1D}} \in \mathbb{R}^{n\times d_v}$, where $n<thw$ reflects the compression.

To align each visual token $z_{\operatorname{1D}} \in Z_{\operatorname{1D}}$ with the  pretrained LLM vocabulary, we introduce a generator module $\mathcal{G}(\cdot)$ in conjunction with the Gumbel-softmax technique~\cite{jang2016categorical,kusner2016gans}. This generator maps each visual token to a soft categorical distribution on an existing LLM ~\cite{jiang2024mixtral,yang2024qwen2,dubey2024llama} vocabulary $\mathbf{L} \in \mathbb{R}^{N_{l} \times d_{l}}$, containing $N_{l}$ items. Note that the vocabulary remains fixed throughout the optimization process to preserve consistent sematic grounding. 
Specifically, the soft categorical distribution  for each visual token is computed as follows:
\begin{equation}
\alpha_{k}=\frac{\exp\left(\mathcal{G}(z_{\operatorname{1D}})_{k}+g_k\right)}{\sum_{j=1}^{N_l} \exp \left(\mathcal{G}(z_{\operatorname{1D}})_{j}+g_j\right)}, \quad \forall k \in\{1,2, \ldots, N_l\}
\end{equation}
where  $g_k \sim \operatorname{Gumbel}(0,1)$ represents  independently sampled  Gumbel noise, and $\mathcal{G}(\cdot): \mathbb{R}^{1  \times d_{v}} \rightarrow \mathbb{R}^{1 \times N_l}$ produces probability logits across  $N_{l}$ categories. 

The resulting logits $\boldsymbol{\alpha}=\left(\alpha_1, \alpha_2, \ldots, \alpha_{{N}_l}\right)$ thus form  a categorical distribution for each element $z_{\operatorname{1D}}$ over the LLM vocabulary. Each visual token is then embedded into the vocabulary space via 
$\mathbf{v}=\boldsymbol{\alpha} \cdot \mathbf{L}$, 
where $\mathbf{v}$ represents the \emph{visual-vocabulary aligned} embedding. By directly operating in the LLM vocabulary space, our approach bridges visual and linguistic modalities while preserving crucial visual content.

 Subsequently, each element $\mathbf{v}$ is quantized by matching it to the closest code in the visual vocabulary codebook. We employ an  exponentially moving average (EMA) scheme to  update the codebook, consistent with findings that EMA improves convergence and training stability \cite{van2017neural,yan2021videogpt}.

\subsection{Masked Autoregressive Rectified-Flow Decoder}
\label{flow}
Once visual elements are discretized into a feature space aligned with the vocabulary of pretrained LLMs, a critical challenge remains: how to reconstruct the original image with high-fidelity and detailed granularity.
Drawing on recent advances in masked generative modeling, we introduce a transformer encoder-decoder architecture, augmented with a masking strategy, to facilitate flow-matching for high-quality visual reconstruction. 

 We first partition the input image $X$ into $N$ non-overlapping patches, forming a sequence $\left\{x^0, x^{1} \ldots, x^N\right\}$. A masking ratio $\rho \in [0.7, 1.0]$ is then randomly selected (e.g., $\rho=0.7$ masks 70\% of the patches.). To incorporate contextual information from both the umasked patches and  the quantized visual tokens, we concatenate them and feed the combined sequence into a transformer encoder.
 
For the reconstruction of  masked patches, we initialize learnable mask tokens \texttt{[m]}, which are appended to the encoder output and  passed into a transformer decoder. The decoder produces contextual embeddings for masked tokens \texttt{[m]}. These embeddings  serve  as conditioning signals for the velocity network in ~\cref{ode}. Formally, the reconstruction process of masked  patches is formulated as follows:
\begin{equation}
p\left(x^{j}_{0: t} \mid \texttt{[m]} \right)=p\left(x^j_{t}\right) \prod_{i=1}^t p\left(x^j_{(i-1) \cdot \Delta t} \mid z^j_{i \cdot \Delta t}, \texttt{[m]},t\right)
\end{equation}
where  $j \in \Omega$ denotes the set of masked patches. The notation $p\left(x^{j}_{0: t} \mid \texttt{[m]} \right)$ represents that recovering target visual content from noised inputs ${x}^j_t = \epsilon$, conditioned on masked patches \texttt{[m]}; $\Delta t$ is the time-step size. Since the masked patches  \texttt{[m]}  already capture semantically rich  visual  representations through masked image modeling, we find that a lightweight MLP suffices to achieve high-fidelity visual reconstruction. This design simplifies the search for an optimal probabilistic path in the flow-matching process.\\
\textbf{Rectified-flow Sampling Stage.} At sampling time, we employ an autogressive ``next set-of-tokens prediction" scheme akin to  MAR~\cite{li2024autoregressive},  progressively decreasing the masking ratio from 1.0 to 0 via a  cosine schedule. 
At the initial iteration, with a masking ratio of 1.0, only the quantized visual tokens are fed  into the transformer encoder-decoder, and a subset of tokens is randomly selected for reconstruction. In subsequent iterations,  the newly reconstructed patches are concatenated with previously reconstructed patches and the quantized visual tokens, then fed back into the encoder-decoder,  continuing to reconstruct the next subset of masked patches. 
The overall sampling process follows:
\begin{equation}
\resizebox{\linewidth}{!}{$
p\left(\mathbf{x}^{1}, \cdots, \mathbf{x}^K\right)=\left\{\begin{array}{l}
\prod\limits_{k}^{K} p\left(\mathbf{x}^k \mid \mathbf{V}^2 \right), \quad\quad\quad\quad\quad\ k=1 \\
\prod\limits_{k}^{K} p\left(\mathbf{x}^k \mid \mathbf{V}^2, \mathbf{x}^{1}, \cdots, \mathbf{x}^{k-1}\right), k>1
\end{array}\right.$}
\end{equation}
Here, $\mathbf{x}^k=\left\{x^i, x^{i+1} \ldots, x^j\right\}$ denotes the patches 
 reconstructed at the k-th iteration, and $\cup_k \mathbf{x}^k=\left\{x^0, x^{1} \ldots, x^N\right\}$ spans the full set of patches.
Subsequently, the masked tokens produced by the decoder serve as conditioning signals for a lightweight velocity model to reconstruct the corresponding patches from noised samples.
Classifier-free guidance is also incorporated during sampling, by interpolating between $\psi_{\Theta}\left(z_t, t,\mathbf{V}^2\right)$ and its unconditional counterpart $\psi_{\Theta}\left(z_t, t,\mathbf{D}\right)$ via a scaling factor $w$, 
\begin{equation}
\label{cfg}
    \tilde{\psi}_{\Theta}\left(z_t, t,\mathbf{V}^2\right) = \omega {\psi}_{\Theta}\left(z_t, t,\mathbf{V}^2\right) + (1 - \omega) {\psi}_{\Theta}\left(z_t, t,\mathbf{D}\right)
\end{equation}
where $\mathbf{D}$ is a learnable dummy token under unconditional sampling.
Overall,  \cref{alg:duplicate_items} summarizes the sampling procedure of  masked autoregressive rectified-flow decoder.
\begin{algorithm}[]
\caption{Sampling Procedure of the Decoder}
\label{alg:duplicate_items}
\begin{algorithmic}[1]
\REQUIRE
$\mathbf{V}^2$ - Token sequence from  the visual vocabulary resampler, $steps$ - Autoregressive iteration steps \\
\ENSURE $\cup_k \mathbf{x}^k$ - Complete set of  reconstructed  patches
\STATE $set \gets \text{new set}()$
\FOR{$k \gets 0$ \TO $steps$}
    \STATE Concatenate $\mathbf{V}^2$ with  $\left\{\mathbf{x}^0,\ldots,\mathbf{x}^{k-1}\right\}$ and input into the transformer encoder.
    \STATE Sample a masking ratio for iteration $k$ from a cosine schedule in $[1.0 \rightarrow 0]$.
    \STATE Reconstruct  newly masked patches using classifier-free guidance  in \cref{cfg}.
    \STATE $set.update(\mathbf{x}^k)$
\ENDFOR
\RETURN $\cup_k \mathbf{x}^k$ 
\end{algorithmic}
\end{algorithm}

\subsection{Equipped with LLM for Visual Generation}
\begin{figure}[]
  \centering
   \includegraphics[width=0.9\linewidth]{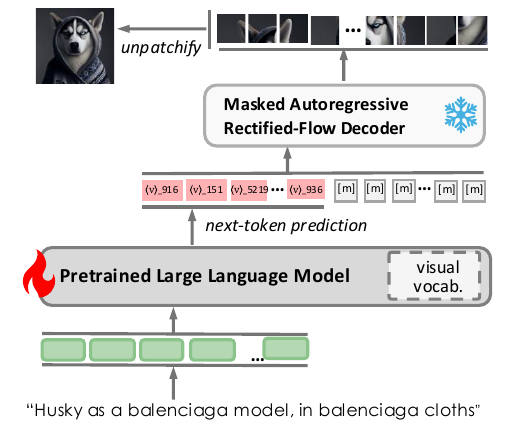}

   \caption{ Pipline for  integrating 
V\textsuperscript{2}Flow tokenizer with pretrained LLMs for autoregressive visual generation.}
   \label{fig:overview_ar}
\end{figure}

\cref{fig:overview_ar} illustrates the integration pipeline of our  V\textsuperscript{2}Flow tokenizer with LLMs for autoregressive visual generation. To facilitate seamless integration, we expand the existing LLM vocabulary with a set of visual-specific tokens, denoted as \texttt{<v>\_0}, \texttt{<v>\_1}, \texttt{<v>\_2},..., \texttt{<v>\_{No}}, initialized directly using the codebook embeddings from our V\textsuperscript{2}Flow tokenizer. Here,  the notation \texttt{No} is the size of the  visual vocabulary codebook. (See \cref{sec3.1}.)

During training, we formulate single-turn conversation data comprising text-image pairs, where the text prompt serves as the instruction, and the discrete visual tokens  constitute the predictive target response. The LLM is optimized using an autoregressive training objective~\cite{touvron2023llama,liu2023visual}, learning to predict the visual token sequence and determine appropriate stopping criteria.

At inference time, the pretrained LLM generates visual tokens autoregressively based on the input prompt until reaching a stopping token. Subsequently, the generated discretized visual tokens are fed into the V\textsuperscript{2}Flow decoder, where rectified-flow sampling reconstructs the original high-quality image, illustrated in \cref{alg:duplicate_items}.


\section{Experiments}
\label{sec:experiments}

\subsection{Experimental Settings}

\textbf{Datasets.} For training the image tokenizer, we exclusively utilize the ImageNet training split~\cite{deng2009imagenet}. The reconstruction performance is then evaluated on the ImageNet-1K test subset, which contains 100,000 images. Evaluations are conducted at two distinct resolutions: $256 \times 256$ and $512 \times 512$ pixels, enabling a comprehensive analysis of the model's performance across varying image scales.

For the autoregressive text-conditioned image generation, the training dataset comprises approximately 6M image-text pairs from JourneyDB~\cite{sun2023journeydb} and LAION-HD~\cite{schuhmann2022laion}. All images are center-cropped and resized proportionally to a resolution of $512 \times 512$ pixels.\\
\textbf{Implementation Details.} To achieve autoregressive text-conditioned image generation, we employ the LLaMA2-7B language decoder~\cite{touvron2023llama} as our visual generative backbone. 
At the training stage, we empirically select a  global batch size of 1,024 and a learning rate of 2e-5, training the model for 10 epochs. Consistent with  LLaVA~\cite{liu2023visual}, we use the Adam optimizer without weight decay, applying  a cosine learning rate schedule with a warm-up ratio of 3\%. For implementation details of training V\textsuperscript{2}Flow tokenizer , please refer to the \textbf{supplementary material.}

\begin{table}[!t]
\centering
\renewcommand{\arraystretch}{1.25}
\resizebox{\linewidth}{!}{%
\begin{tabular}{p{3.1cm}>{\centering\arraybackslash}p{1cm}c|>{\centering\arraybackslash}p{0.8cm}>{\centering\arraybackslash}p{0.8cm}>{\centering\arraybackslash}p{0.8cm}}
\toprule
\multirow{2}{*}{\textbf{VQ-Tokenizer}} & 
\multirow{2}{*}{\textbf{Tokens}} & 
\multirow{2}{*}{\textbf{Codebook}} &
\multicolumn{3}{c}{ \textbf{ImageNet-1K}} \\
\cline{4-6}
  &  &   & 
 \textbf{PSNR}~$\uparrow$ & \textbf{SSIM}~$\uparrow$ &  \textbf{LPIPS}~$\downarrow$ \\
\midrule
\multicolumn{3}{l}{\textbf{Resolution $256 \times 256$:}}\\
\midrule
 Taming-VQGAN~\cite{esser2021taming}  & $16\times16$ & 16,384 & 19.68 & 0.50  & 0.29 \\
 IBQ~\cite{shi2024taming}  & $16\times16$ & 16,384 & 21.25 & 0.58  & 0.23 \\
 LlamaGen~\cite{sun2024autoregressive} & $16\times16$ & 16,384  & 20.86 & 0.57  & 0.24  \\
 TiTok~\cite{yu2024image}  & 64 & 4,096  & 18.65  & 0.44  & 0.34  \\
 TiTok~\cite{yu2024image}  & 128  & 4,096  & 19.97 & 0.51  & 0.28 \\
 TiTok~\cite{yu2024image}  & 256  & 4,096  & 21.44 & 0.59  & 0.24 \\
 Open-MAGVIT2~\cite{luo2024open} & $16\times16$ & 262,144  & 21.76 & 0.60 & 0.21 \\
 Cosmos-DI~\cite{agarwal2025cosmos}  & $16\times16$ & 64,000  & 20.38 & 0.52  & 0.28   \\
 V\textsuperscript{2}Flow  &   256 & 16,384 &  \textbf{22.37} & \textbf{0.65}  & \textbf{0.20}   \\
\midrule
\multicolumn{3}{l}{\textbf{Resolution $512 \times 512$:}}\\
\midrule
Taming-VQGAN ~\cite{esser2021taming}
 & $32\times32$ & 16,384 & 21.73 & 0.58 & 0.29 \\
 IBQ~\cite{shi2024taming}  & $32\times32$ & 16,384 & 23.33 & 0.64  & 0.23 \\
 LlamaGen~\cite{sun2024autoregressive}   & $32\times32$ & 16,384  &  23.02 & 0.64 &  0.24 \\
 Open-MAGVIT2~\cite{luo2024open} & $32\times32$ & 262,144  & \textbf{23.80} & 0.65 & 0.22 \\
 Cosmos-DI~\cite{agarwal2025cosmos}  & $32\times32$ & 64,000 & 22.57 & 0.61 & 0.27 \\
 V\textsuperscript{2}Flow  & 1024 & 16,384 & \underline{23.28} & \textbf{0.65} & \textbf{0.22} \\
\bottomrule
\end{tabular}%
}
\caption{Quantitative comparison with prior arts VQ-tokenizers at resolution $256 \times256$ and $512 \times 512$ on ImageNet-1K test subset. Evaluation metrics including PSNR~\cite{hore2010image},SSIM~\cite{zhang2018unreasonable} and LPIPS~\cite{wang2004image} are reported.}
\label{tab:di_tokenizer_evaluation}
\end{table}
\begin{table}[]
\centering
\renewcommand{\arraystretch}{1.2}
\resizebox{\linewidth}{!}{
\begin{tabular}{ccccc}
\toprule
\textbf{ImageNet 256$\times$256} & \textbf{Autoencoder} &  \textbf{PSNR} $\uparrow$ & \textbf{SSIM} $\uparrow$ & \textbf{LPIPS} $\downarrow$ \\
\midrule
\rowcolors{2}{gray!20}{white}
\multirow{2}{*}{64 tokens}   &  TiTok~\cite{yu2024image}  & 18.65  & 0.44  & 0.34  \\
                         & V\textsuperscript{2}Flow    &  \textbf{20.41} & \textbf{0.52} & \textbf{0.26}     \\
\midrule
\multirow{2}{*}{128 tokens}   &  TiTok~\cite{yu2024image}  & 19.97 & 0.51  & 0.28  \\
                         & V\textsuperscript{2}Flow & \textbf{22.08}    & \textbf{0.63} & \textbf{0.21}      \\
\midrule
\multirow{2}{*}{256 tokens}   &  TiTok~\cite{yu2024image}  & 21.44 & 0.59  & 0.24  \\
                         & V\textsuperscript{2}Flow  & \textbf{22.37} & \textbf{0.65}  & \textbf{0.20} \\
\bottomrule
\end{tabular}
}
\caption{Comparison of reconstruction quality with the 1D  TiTok tokenizer~\cite{yu2024image}  on ImageNet 256$\times$256 at various token lengths.}
\label{tab:titok}
\end{table}
\begin{table}[]
\centering
\renewcommand{\arraystretch}{1.2}
\resizebox{\linewidth}{!}{
\begin{tabular}{cccccc}
\toprule
\textbf{DDIM} & \textbf{Rectified-Flow} & \textbf{CFG} &  \textbf{PSNR} $\uparrow$ & \textbf{SSIM} $\uparrow$ & \textbf{LPIPS} $\downarrow$ \\
\midrule
\rowcolors{2}{gray!20}{white}
& \checkmark &    & 21.44 & 0.59  & 0.24  \\
                         & \checkmark  &    \checkmark        &   \textbf{22.37} & \textbf{0.65} & \textbf{0.20} \\
 \checkmark  &    & \checkmark    & 22.23  & 0.64  & 0.20 \\
\bottomrule
\end{tabular}
}
\caption{Ablation study of sampling strategy including classifier-free guidance (CFG) and different sampling strategy in the masked autoregressive decoder.}
\label{tab:autoencoder_comparison}
\end{table}

\begin{figure*}[]
  \centering
   \includegraphics[width=\textwidth]{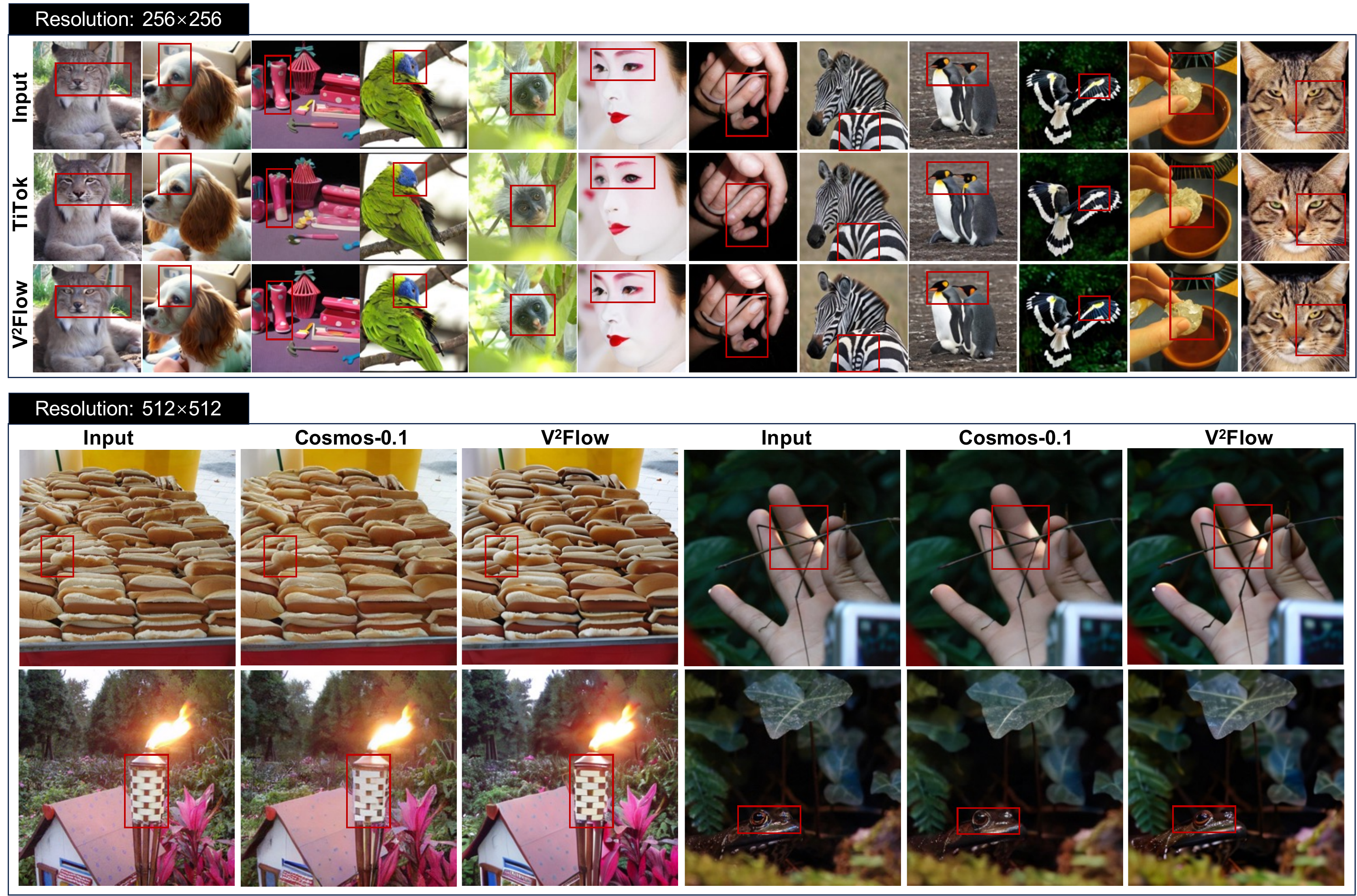}

   \caption{ Qualitative comparisons of reconstruction quality on the ImageNet-1K test subset, comparing V\textsuperscript{2}Flow against  TiTok~\cite{yu2024image} at resolution 256×256 and the CosMos-Discrete~\cite{agarwal2025cosmos} at resolution 512×512. For resolution $256 \times 256$, both  V\textsuperscript{2}Flow and TiTok ~\cite{yu2024image}compress the input image  into an one-dimensional sequence of 256 tokens, yet V\textsuperscript{2}Flow reconstructs  images with finer details. At resolution $512 \times 512$, compared to CosMos which compresses input images into 2D grid latents, V\textsuperscript{2}Flow still achieves superior reconstruction quality.
   }
   \label{fig:quality}
\end{figure*}

\begin{figure*}[ht]
  \centering
   \includegraphics[width=\textwidth]{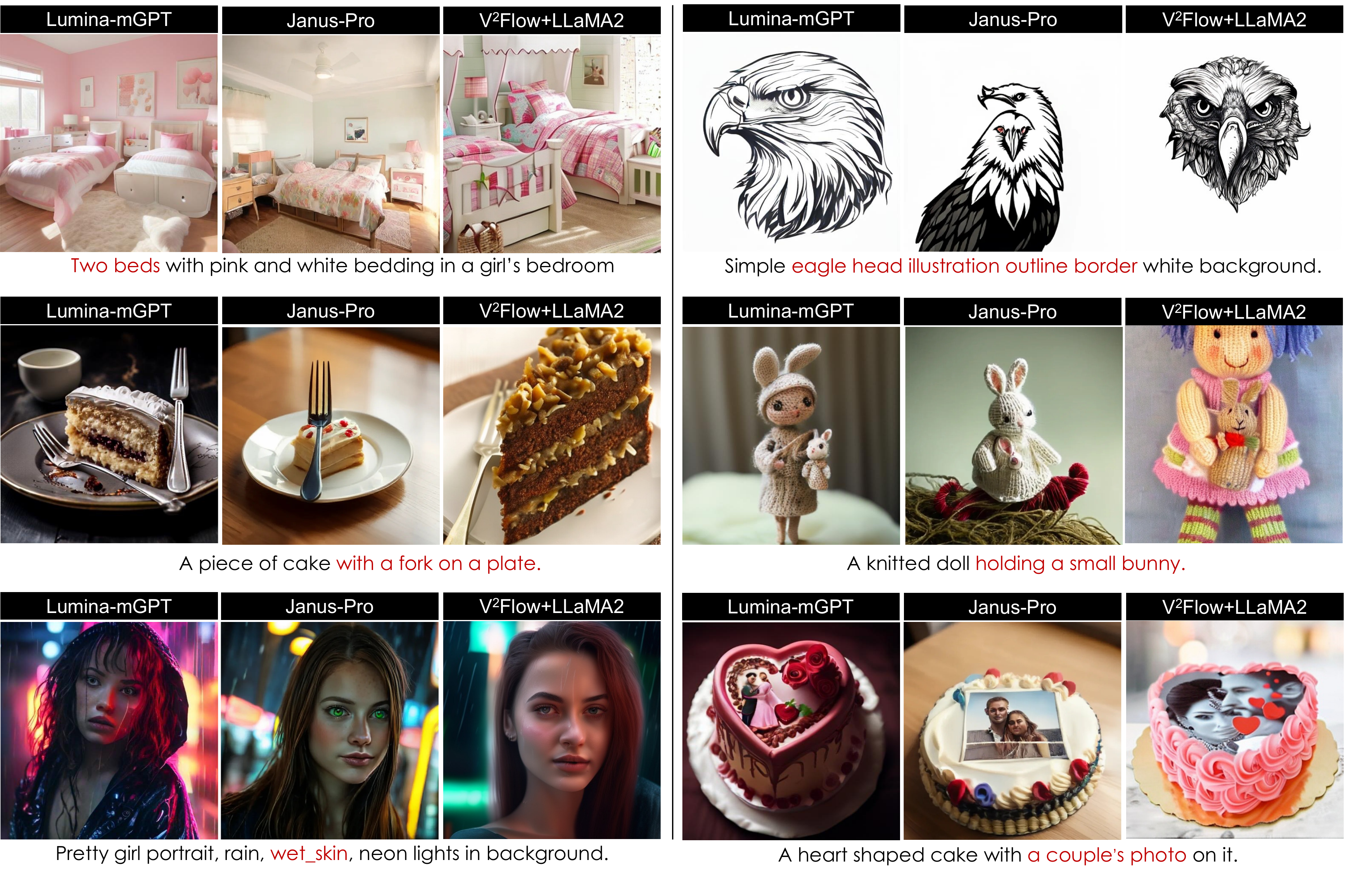}

   \caption{ Qualitative results of text-conditioned image generation. We compare our approach against recent state-of-the-art autoregressive models, including Janus-Pro-7B~\cite{chen2025janus} and Lumina-mGPT-7B~\cite{liu2024lumina}. All images are generated at a resolution of  512×512.
   }
   \label{fig:quality}
\end{figure*}

\subsection{Results on Reconstruction Quality}
\textbf{Quantitative Results.} 
\cref{tab:di_tokenizer_evaluation} summarizes the average quantitative metrics of  our tokenizer on ImageNet-1K test subset of resolution $256 \times 256$ and $512 \times 512$.
For ImageNet $256 \times 256$ results, ours tokenizer achieves state-of-the-art performance in all the metrics compared to prior arts. More importantly, even with a lower codebook size compared to the competitive Open-MAGVIT2~\cite{luo2024open} ( from 16,384 to 262,144), our tokenizer improves over  Open-MAGVIT2 by a large margin.
For ImageNet $512 \times 512$ results, our tokenizer consistently achieves the competitive results compared to prior arts. These quantitative results confirm that our tokenizer can effectively  represent visual content within the vocabulary space of pretrained LLMs, and accurately reconstruct original images while preserving pivotal details.\\
\textbf{Qualitative Analysis.}
In \cref{fig:quality}, we compare reconstruction results on the ImageNet-1K test subset, evaluating our tokenizer against the competitve 1D tokenizer TiTok~\cite{yu2024image} at a $256 \times 256$ resolution and  the 2D CosMos tokenizer~\cite{agarwal2025cosmos} at a $512 \times 512$ resolutions, respectively. The visualization clearly indicates that our tokenizer more faithfully restore critical visual details that are lost in the reconstruction results of the competing tokenizers at both resolutions. These findings underscore the effectiveness of our tokenizer in preserving fine-grained visual content, highlighting its superior reconstruction fidelity.\\
\textbf{Flexibility in Tokenization Length.}
\label{sec:experiments.rec}
 A key advantage of our proposed masked autoregressive rectified-flow decoder is its ability to tokenize visual content at varying sequence lengths while preserving robust reconstruction performance. In ~\cref{tab:titok}, we compare the reconstruction results across different token lengths against the state-of-the-art TiTok tokenizer~\cite{yu2024image} on the ImageNet-1k test set at a resolution of 256×256. The experimental findings reveal that our method consistently outperforms TiTok across three different token sequence lengths. Notably, even with only 128 tokens, our approach surpasses TiTok’s reconstruction performance at 256 tokens, underscoring the efficiency and flexibility of the rectified-flow decoder.\\
\textbf{Ablation study of sampling strategy.}  In ~\cref{tab:autoencoder_comparison} , we first  analyze the effect of  classifier-free guidance during the masked autoregressive sampling stage. Quantitative results show  that integrating classifier-free guidance  achieves gains of   0.93 and 0.06 in PSNR and SSIM metrics. In addition, we examine the sensitivity of our tokenizer to different 
probability distribution modeling paradigms. When substituting the rectified-flow formulation  with DDIM~\cite{song2020denoising}, our approach maintains competitive reconstruction quality. This  confirms  the robustness of the  sampling strategy across different probability distribution modeling paradigms.


\subsection{Results on Text-to-Image Generation}

\begin{figure}[!t]
  \centering
   \includegraphics[width=\linewidth]{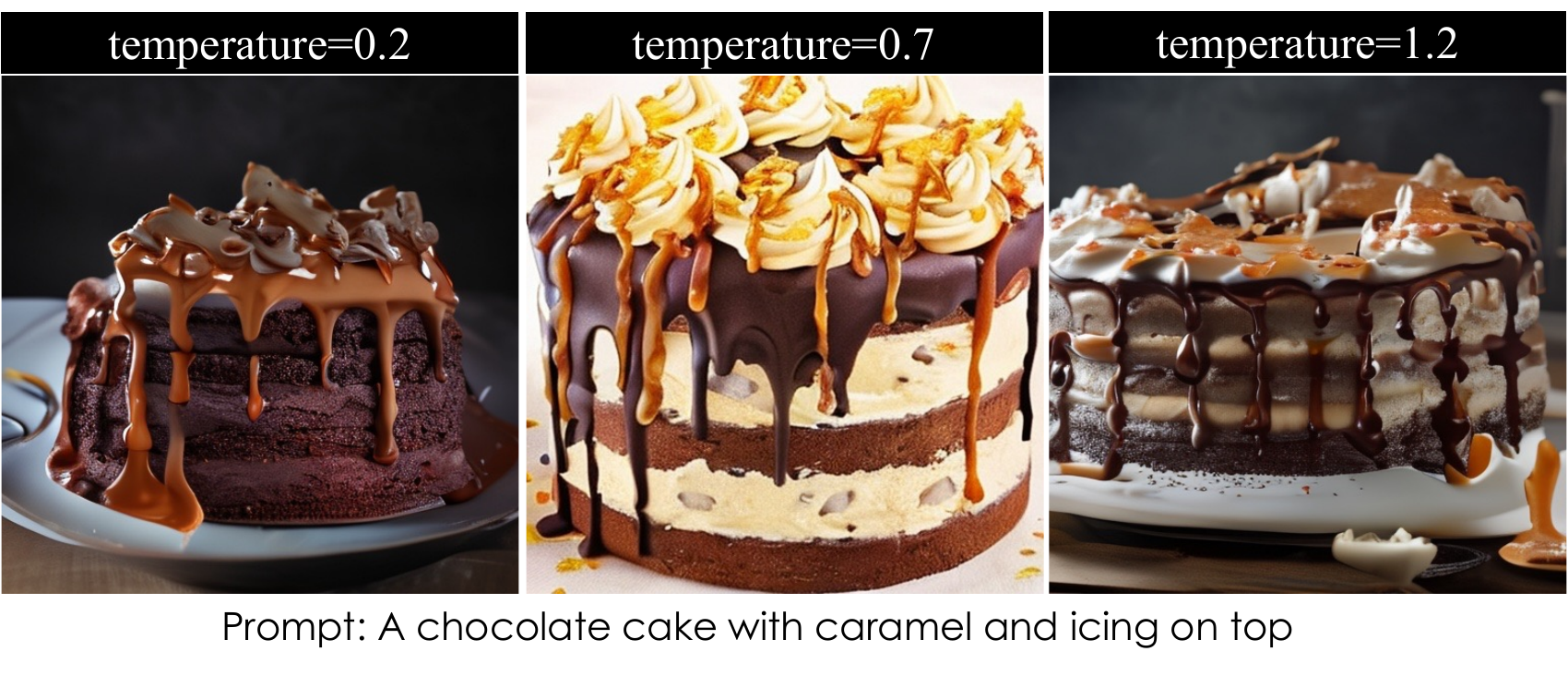}

   \caption{Analysis of the temperature hyperparameter during the autoregressive sampling process.
   }
   \label{fig:param_analysis}
\end{figure}

\textbf{Qualitative Analysis.}
\cref{fig:quality} provides a qualitative comparison of our text-to-image generation results against two state-of-the-art autoregressive models, Janus-Pro-7B~\cite{chen2025janus} and Lumina-mGPT-7B~\cite{liu2024lumina}. The visualization illustrates that our V\textsuperscript{2}Flow tokenizer, integrated with the LLaMA2-7B language model~\cite{touvron2023llama}, more precisely captures semantic details from textual prompts. Consequently, our method achieves competitive generation quality, consistently producing coherent and contextually relevant images.\\
\textbf{Temperature of  Autoregressive Sampling.}
In ~\cref{fig:param_analysis}, we present an analysis of the temperature hyperparameter during the autoregressive sampling process. We observe that appropriately tuned temperatures enhance diversity and introduce beneficial variability in the generated images. However, excessively high temperatures can compromise image coherence, resulting in semantically inconsistent or fragmented outcomes.

\section{Related Works}
\label{sec:related}
\textbf{Visual Quantization.} VQ-VAE ~\cite{van2017neural} stands as a pivotal work in the field of image quantization~\cite{lee2022autoregressive,yu2021vector,peng2022beit}. VQ-GAN ~\cite{yu2021vector} further refines this apporach by incorporating adversarial and perceptual losses to capture more precise visual elements. Subsequent methods, including
RQ-VAE~\cite{lee2022autoregressive} and MoVQ, ~\cite{zheng2022movq} explore  multiple vector quantization steps per latent
embedding. 
MAGVIT-v2 ~\cite{yu2023language} and FSQ ~\cite{mentzer2023finite} introduce lookup-free quantization strategies, leading to large visual codebooks and expressive representations. TiTok~\cite{yu2024image} adopts a  masked transformer encoder-decoder to tokenize images at resolution 
 $256 \times 256$ into an one-dimensional sequence of 32 discrete tokens only. 
 
 Despite advances, the latent distributions of quantized visual tokens diverge significantly from those of text. The disparities between  two modalities impose considerable challenges for autoregressive modeling.
Although recent efforts~\cite{zhu2024beyond,yu2024spae}  
address this by utilizing  a LLM's fixed vocabulary as the visual codebook, aligning visual and linguistic modalities more directly,  the  resulting tokens typically preserve the intrisic two-dimensional structure of images. Consequently,  autogressive models  must predict visual tokens in a line-by-line manner, deviating from the   one-dimensional text-processing approach used by existing LLMs.\\
\textbf{Autogressive Visual Generation.}
Most existing  autoregressive visual generation models~\cite{liu2024lumina,chen2025janus,team2405chameleon,zhou2024transfusion,sun2024autoregressive,dubey2024llama,tian2024visual} primarily focus on a sequential pixel-by-pixel process.  Chameleon~\cite{team2405chameleon} simultaneously addresses  image captioning and generation within a unified Transformer framework. 
Janus~\cite{chen2025janus} decouples visual encoding
into separate pathways yet employs a single transformer  for
multimodal understanding and generation. 
Lumina-mGPT~\cite{liu2024lumina} captures extensive multimodal capabilities by applying a  next-token prediction objective over interleaved text-image sequences. Transfusion~\cite{zhou2024transfusion} integrates  next-token prediction for text with 
diffusion-based generation for images, unifying discrete and continuous modalities in one system. LlamaGen~\cite{sun2024autoregressive}, built on vanilla autoregressive models, deliberately avoids  visual  inductive biases, instead advancing  image generation  through proper scaling. However,  a fundamental challenge remains: these approaches typically convert two-dimensional images into one-dimensional sequences via a raster-scan ordering, limiting their ability to capture global
structure. VAR~\cite{tian2024visual} attempts to address this concern by reframing  autoregressive visual generation as a coarse-to-fine ``next-scale” or ``next-resolution” prediction process. However, it  remains susceptible to  error accumulation when predicting multiple tokens in parallel.

In contrast,  our   V\textsuperscript{2}Flow tokenizer compresses visual content into discrete token sequences that support  superior reconstruction  performance while maintaining both semantic and structural alignment with text representations in existing LLMs. This alignment facilitates  autoregressive visual generation directly on top of existing LLMs.\\
\textbf{Comparison between V\textsuperscript{2}Flow and MAR.} Our proposed Masked Autoregressive Rectified-Flow  decoder within V\textsuperscript{2}Flow shares conceptual ground with MAR~\cite{li2024autoregressive}. However, beyond implementation differences, e.g., our adoption of a more  efficient Rectified-Flow sampling strategy instead of MAR's vanilla DDPM, we introduce a crucial conceptual advancement by capturing cross-modal mappings explicitly. We leverage masked image modeling to learn powerful contextual priors, facilitating detailed visual reconstructions guided by compact textual conditioning signals. This enables more effective integration of visual and textual modalities and paves the way toward a unified autoregressive framework for multimodal generation.

\section{Conclusion}
In this work, we introduce V\textsuperscript{2}Flow tokenizer,  a vector-quantized visual  tokenizer  with two core designs. First, we propose a  visual vocabulary resampler transforming  visual content into a compact token sequence expressed within pretrained LLM's vocabulary space. This facilitates  seamless integration into pretrained LLMs for autoregressive visual generation.  Second, we develop masked autoregressive rectified-flow decoder in tandem with ``next set-of-tokens" sampling strategy. This facilitates visual tokenization at varying sequence lengths while preserving robust and competitive reconstruction performance. Experimental results show that our V\textsuperscript{2}Flow achieves competitive reconstruction performance compared to mainstream VQ-based tokenizers  and effectively advances autoregressive visual generation.

{
    \small
    \bibliographystyle{ieeenat_fullname}
    \bibliography{main}

\begin{thebibliography}{61}
\providecommand{\natexlab}[1]{#1}
\providecommand{\url}[1]{\texttt{#1}}
\expandafter\ifx\csname urlstyle\endcsname\relax
  \providecommand{\doi}[1]{doi: #1}\else
  \providecommand{\doi}{doi: \begingroup \urlstyle{rm}\Url}\fi

\bibitem[Agarwal et~al.(2025)Agarwal, Ali, Bala, Balaji, Barker, Cai, Chattopadhyay, Chen, Cui, Ding, et~al.]{agarwal2025cosmos}
Niket Agarwal, Arslan Ali, Maciej Bala, Yogesh Balaji, Erik Barker, Tiffany Cai, Prithvijit Chattopadhyay, Yongxin Chen, Yin Cui, Yifan Ding, et~al.
\newblock Cosmos world foundation model platform for physical ai.
\newblock \emph{arXiv preprint arXiv:2501.03575}, 2025.

\bibitem[Bai et~al.(2023)Bai, Bai, Chu, Cui, Dang, Deng, Fan, Ge, Han, Huang, et~al.]{bai2023qwen}
Jinze Bai, Shuai Bai, Yunfei Chu, Zeyu Cui, Kai Dang, Xiaodong Deng, Yang Fan, Wenbin Ge, Yu Han, Fei Huang, et~al.
\newblock Qwen technical report.
\newblock \emph{arXiv preprint arXiv:2309.16609}, 2023.

\bibitem[Cha et~al.(2024)Cha, Kang, Mun, and Roh]{cha2024honeybee}
Junbum Cha, Wooyoung Kang, Jonghwan Mun, and Byungseok Roh.
\newblock Honeybee: Locality-enhanced projector for multimodal llm.
\newblock In \emph{Proceedings of the IEEE/CVF Conference on Computer Vision and Pattern Recognition}, pages 13817--13827, 2024.

\bibitem[Chen et~al.(2025)Chen, Wu, Liu, Pan, Liu, Xie, Yu, and Ruan]{chen2025janus}
Xiaokang Chen, Zhiyu Wu, Xingchao Liu, Zizheng Pan, Wen Liu, Zhenda Xie, Xingkai Yu, and Chong Ruan.
\newblock Janus-pro: Unified multimodal understanding and generation with data and model scaling.
\newblock \emph{arXiv preprint arXiv:2501.17811}, 2025.

\bibitem[Cheng et~al.(2024)Cheng, Leng, Zhang, Xin, Li, Chen, Zhu, Zhang, Luo, Zhao, et~al.]{cheng2024videollama}
Zesen Cheng, Sicong Leng, Hang Zhang, Yifei Xin, Xin Li, Guanzheng Chen, Yongxin Zhu, Wenqi Zhang, Ziyang Luo, Deli Zhao, et~al.
\newblock Videollama 2: Advancing spatial-temporal modeling and audio understanding in video-llms.
\newblock \emph{arXiv preprint arXiv:2406.07476}, 2024.

\bibitem[Deng et~al.(2009)Deng, Dong, Socher, Li, Li, and Fei-Fei]{deng2009imagenet}
Jia Deng, Wei Dong, Richard Socher, Li-Jia Li, Kai Li, and Li Fei-Fei.
\newblock Imagenet: A large-scale hierarchical image database.
\newblock In \emph{2009 IEEE conference on computer vision and pattern recognition}, pages 248--255. Ieee, 2009.

\bibitem[Dong et~al.(2023)Dong, Han, Peng, Qi, Ge, Yang, Zhao, Sun, Zhou, Wei, et~al.]{dong2023dreamllm}
Runpei Dong, Chunrui Han, Yuang Peng, Zekun Qi, Zheng Ge, Jinrong Yang, Liang Zhao, Jianjian Sun, Hongyu Zhou, Haoran Wei, et~al.
\newblock Dreamllm: Synergistic multimodal comprehension and creation.
\newblock \emph{arXiv preprint arXiv:2309.11499}, 2023.

\bibitem[Dubey et~al.(2024)Dubey, Jauhri, Pandey, Kadian, Al-Dahle, Letman, Mathur, Schelten, Yang, Fan, et~al.]{dubey2024llama}
Abhimanyu Dubey, Abhinav Jauhri, Abhinav Pandey, Abhishek Kadian, Ahmad Al-Dahle, Aiesha Letman, Akhil Mathur, Alan Schelten, Amy Yang, Angela Fan, et~al.
\newblock The llama 3 herd of models.
\newblock \emph{arXiv preprint arXiv:2407.21783}, 2024.

\bibitem[Esser et~al.(2021)Esser, Rombach, and Ommer]{esser2021taming}
Patrick Esser, Robin Rombach, and Bjorn Ommer.
\newblock Taming transformers for high-resolution image synthesis.
\newblock In \emph{Proceedings of the IEEE/CVF conference on computer vision and pattern recognition}, pages 12873--12883, 2021.

\bibitem[Esser et~al.(2024)Esser, Kulal, Blattmann, Entezari, M{\"u}ller, Saini, Levi, Lorenz, Sauer, Boesel, et~al.]{esser2024scaling}
Patrick Esser, Sumith Kulal, Andreas Blattmann, Rahim Entezari, Jonas M{\"u}ller, Harry Saini, Yam Levi, Dominik Lorenz, Axel Sauer, Frederic Boesel, et~al.
\newblock Scaling rectified flow transformers for high-resolution image synthesis.
\newblock In \emph{Forty-first international conference on machine learning}, 2024.

\bibitem[Fang et~al.(2024)Fang, Duan, Wang, Li, Tian, Zeng, Zhao, Dai, Li, and Liu]{fang2024puma}
Rongyao Fang, Chengqi Duan, Kun Wang, Hao Li, Hao Tian, Xingyu Zeng, Rui Zhao, Jifeng Dai, Hongsheng Li, and Xihui Liu.
\newblock Puma: Empowering unified mllm with multi-granular visual generation.
\newblock \emph{arXiv preprint arXiv:2410.13861}, 2024.

\bibitem[Ge et~al.(2024)Ge, Zhao, Zhu, Ge, Yi, Song, Li, Ding, and Shan]{ge2024seed}
Yuying Ge, Sijie Zhao, Jinguo Zhu, Yixiao Ge, Kun Yi, Lin Song, Chen Li, Xiaohan Ding, and Ying Shan.
\newblock Seed-x: Multimodal models with unified multi-granularity comprehension and generation.
\newblock \emph{arXiv preprint arXiv:2404.14396}, 2024.

\bibitem[He et~al.(2022)He, Chen, Xie, Li, Doll{\'a}r, and Girshick]{he2022masked}
Kaiming He, Xinlei Chen, Saining Xie, Yanghao Li, Piotr Doll{\'a}r, and Ross Girshick.
\newblock Masked autoencoders are scalable vision learners.
\newblock In \emph{Proceedings of the IEEE/CVF conference on computer vision and pattern recognition}, pages 16000--16009, 2022.

\bibitem[Ho et~al.(2020)Ho, Jain, and Abbeel]{ho2020denoising}
Jonathan Ho, Ajay Jain, and Pieter Abbeel.
\newblock Denoising diffusion probabilistic models.
\newblock \emph{Advances in neural information processing systems}, 33:\penalty0 6840--6851, 2020.

\bibitem[Hore and Ziou(2010)]{hore2010image}
Alain Hore and Djemel Ziou.
\newblock Image quality metrics: Psnr vs. ssim.
\newblock In \emph{2010 20th international conference on pattern recognition}, pages 2366--2369. IEEE, 2010.

\bibitem[Jang et~al.(2016)Jang, Gu, and Poole]{jang2016categorical}
Eric Jang, Shixiang Gu, and Ben Poole.
\newblock Categorical reparameterization with gumbel-softmax.
\newblock \emph{arXiv preprint arXiv:1611.01144}, 2016.

\bibitem[Jiang et~al.(2024)Jiang, Sablayrolles, Roux, Mensch, Savary, Bamford, Chaplot, Casas, Hanna, Bressand, et~al.]{jiang2024mixtral}
Albert~Q Jiang, Alexandre Sablayrolles, Antoine Roux, Arthur Mensch, Blanche Savary, Chris Bamford, Devendra~Singh Chaplot, Diego de~las Casas, Emma~Bou Hanna, Florian Bressand, et~al.
\newblock Mixtral of experts.
\newblock \emph{arXiv preprint arXiv:2401.04088}, 2024.

\bibitem[Kingma(2013)]{kingma2013auto}
Diederik~P Kingma.
\newblock Auto-encoding variational bayes.
\newblock \emph{arXiv preprint arXiv:1312.6114}, 2013.

\bibitem[Kondratyuk et~al.(2023)Kondratyuk, Yu, Gu, Lezama, Huang, Schindler, Hornung, Birodkar, Yan, Chiu, et~al.]{kondratyuk2023videopoet}
Dan Kondratyuk, Lijun Yu, Xiuye Gu, Jos{\'e} Lezama, Jonathan Huang, Grant Schindler, Rachel Hornung, Vighnesh Birodkar, Jimmy Yan, Ming-Chang Chiu, et~al.
\newblock Videopoet: A large language model for zero-shot video generation.
\newblock \emph{arXiv preprint arXiv:2312.14125}, 2023.

\bibitem[Kusner and Hern{\'a}ndez-Lobato(2016)]{kusner2016gans}
Matt~J Kusner and Jos{\'e}~Miguel Hern{\'a}ndez-Lobato.
\newblock Gans for sequences of discrete elements with the gumbel-softmax distribution.
\newblock \emph{arXiv preprint arXiv:1611.04051}, 2016.

\bibitem[Lee et~al.(2022)Lee, Kim, Kim, Cho, and Han]{lee2022autoregressive}
Doyup Lee, Chiheon Kim, Saehoon Kim, Minsu Cho, and Wook-Shin Han.
\newblock Autoregressive image generation using residual quantization.
\newblock In \emph{Proceedings of the IEEE/CVF Conference on Computer Vision and Pattern Recognition}, pages 11523--11532, 2022.

\bibitem[Li et~al.(2024)Li, Tian, Li, Deng, and He]{li2024autoregressive}
Tianhong Li, Yonglong Tian, He Li, Mingyang Deng, and Kaiming He.
\newblock Autoregressive image generation without vector quantization.
\newblock \emph{Advances in Neural Information Processing Systems}, 37:\penalty0 56424--56445, 2024.

\bibitem[Liu et~al.(2024{\natexlab{a}})Liu, Zhao, Zhuo, Lin, Qiao, Li, and Gao]{liu2024lumina}
Dongyang Liu, Shitian Zhao, Le Zhuo, Weifeng Lin, Yu Qiao, Hongsheng Li, and Peng Gao.
\newblock Lumina-mgpt: Illuminate flexible photorealistic text-to-image generation with multimodal generative pretraining.
\newblock \emph{arXiv preprint arXiv:2408.02657}, 2024{\natexlab{a}}.

\bibitem[Liu et~al.(2023)Liu, Li, Wu, and Lee]{liu2023visual}
Haotian Liu, Chunyuan Li, Qingyang Wu, and Yong~Jae Lee.
\newblock Visual instruction tuning.
\newblock \emph{Advances in neural information processing systems}, 36:\penalty0 34892--34916, 2023.

\bibitem[Liu et~al.(2024{\natexlab{b}})Liu, Li, Wu, and Lee]{liu2024visual}
Haotian Liu, Chunyuan Li, Qingyang Wu, and Yong~Jae Lee.
\newblock Visual instruction tuning.
\newblock \emph{Advances in neural information processing systems}, 36, 2024{\natexlab{b}}.

\bibitem[Luo et~al.(2024)Luo, Shi, Ge, Yang, Wang, and Shan]{luo2024open}
Zhuoyan Luo, Fengyuan Shi, Yixiao Ge, Yujiu Yang, Limin Wang, and Ying Shan.
\newblock Open-magvit2: An open-source project toward democratizing auto-regressive visual generation.
\newblock \emph{arXiv preprint arXiv:2409.04410}, 2024.

\bibitem[Mentzer et~al.(2023)Mentzer, Minnen, Agustsson, and Tschannen]{mentzer2023finite}
Fabian Mentzer, David Minnen, Eirikur Agustsson, and Michael Tschannen.
\newblock Finite scalar quantization: Vq-vae made simple.
\newblock \emph{arXiv preprint arXiv:2309.15505}, 2023.

\bibitem[Peng et~al.(2022)Peng, Dong, Bao, Ye, and Wei]{peng2022beit}
Zhiliang Peng, Li Dong, Hangbo Bao, Qixiang Ye, and Furu Wei.
\newblock Beit v2: Masked image modeling with vector-quantized visual tokenizers.
\newblock \emph{arXiv preprint arXiv:2208.06366}, 2022.

\bibitem[Radosavovic et~al.(2020)Radosavovic, Kosaraju, Girshick, He, and Doll{\'a}r]{radosavovic2020designing}
Ilija Radosavovic, Raj~Prateek Kosaraju, Ross Girshick, Kaiming He, and Piotr Doll{\'a}r.
\newblock Designing network design spaces.
\newblock In \emph{Proceedings of the IEEE/CVF conference on computer vision and pattern recognition}, pages 10428--10436, 2020.

\bibitem[Rezende et~al.(2014)Rezende, Mohamed, and Wierstra]{rezende2014stochastic}
Danilo~Jimenez Rezende, Shakir Mohamed, and Daan Wierstra.
\newblock Stochastic backpropagation and approximate inference in deep generative models.
\newblock In \emph{International conference on machine learning}, pages 1278--1286. PMLR, 2014.

\bibitem[Rombach et~al.(2022)Rombach, Blattmann, Lorenz, Esser, and Ommer]{rombach2022high}
Robin Rombach, Andreas Blattmann, Dominik Lorenz, Patrick Esser, and Bj{\"o}rn Ommer.
\newblock High-resolution image synthesis with latent diffusion models.
\newblock In \emph{Proceedings of the IEEE/CVF conference on computer vision and pattern recognition}, pages 10684--10695, 2022.

\bibitem[Schuhmann et~al.(2022)Schuhmann, Beaumont, Vencu, Gordon, Wightman, Cherti, Coombes, Katta, Mullis, Wortsman, et~al.]{schuhmann2022laion}
Christoph Schuhmann, Romain Beaumont, Richard Vencu, Cade Gordon, Ross Wightman, Mehdi Cherti, Theo Coombes, Aarush Katta, Clayton Mullis, Mitchell Wortsman, et~al.
\newblock Laion-5b: An open large-scale dataset for training next generation image-text models.
\newblock \emph{Advances in neural information processing systems}, 35:\penalty0 25278--25294, 2022.

\bibitem[Shi et~al.(2024)Shi, Luo, Ge, Yang, Shan, and Wang]{shi2024taming}
Fengyuan Shi, Zhuoyan Luo, Yixiao Ge, Yujiu Yang, Ying Shan, and Limin Wang.
\newblock Taming scalable visual tokenizer for autoregressive image generation.
\newblock \emph{arXiv preprint arXiv:2412.02692}, 2024.

\bibitem[Song et~al.(2020)Song, Meng, and Ermon]{song2020denoising}
Jiaming Song, Chenlin Meng, and Stefano Ermon.
\newblock Denoising diffusion implicit models.
\newblock \emph{arXiv preprint arXiv:2010.02502}, 2020.

\bibitem[Sun et~al.(2023{\natexlab{a}})Sun, Pan, Ge, Li, Duan, Wu, Zhang, Zhou, Qin, Wang, et~al.]{sun2023journeydb}
Keqiang Sun, Junting Pan, Yuying Ge, Hao Li, Haodong Duan, Xiaoshi Wu, Renrui Zhang, Aojun Zhou, Zipeng Qin, Yi Wang, et~al.
\newblock Journeydb: A benchmark for generative image understanding.
\newblock \emph{Advances in neural information processing systems}, 36:\penalty0 49659--49678, 2023{\natexlab{a}}.

\bibitem[Sun et~al.(2024{\natexlab{a}})Sun, Jiang, Chen, Zhang, Peng, Luo, and Yuan]{sun2024autoregressive}
Peize Sun, Yi Jiang, Shoufa Chen, Shilong Zhang, Bingyue Peng, Ping Luo, and Zehuan Yuan.
\newblock Autoregressive model beats diffusion: Llama for scalable image generation.
\newblock \emph{arXiv preprint arXiv:2406.06525}, 2024{\natexlab{a}}.

\bibitem[Sun et~al.(2023{\natexlab{b}})Sun, Yu, Cui, Zhang, Zhang, Wang, Gao, Liu, Huang, and Wang]{sun2023emu}
Quan Sun, Qiying Yu, Yufeng Cui, Fan Zhang, Xiaosong Zhang, Yueze Wang, Hongcheng Gao, Jingjing Liu, Tiejun Huang, and Xinlong Wang.
\newblock Emu: Generative pretraining in multimodality.
\newblock In \emph{The Twelfth International Conference on Learning Representations}, 2023{\natexlab{b}}.

\bibitem[Sun et~al.(2024{\natexlab{b}})Sun, Cui, Zhang, Zhang, Yu, Wang, Rao, Liu, Huang, and Wang]{sun2024generative}
Quan Sun, Yufeng Cui, Xiaosong Zhang, Fan Zhang, Qiying Yu, Yueze Wang, Yongming Rao, Jingjing Liu, Tiejun Huang, and Xinlong Wang.
\newblock Generative multimodal models are in-context learners.
\newblock In \emph{Proceedings of the IEEE/CVF Conference on Computer Vision and Pattern Recognition}, pages 14398--14409, 2024{\natexlab{b}}.

\bibitem[Team()]{team2405chameleon}
Chameleon Team.
\newblock Chameleon: Mixed-modal early-fusion foundation models, 2024.
\newblock \emph{URL https://arxiv. org/abs/2405.09818}, 9.

\bibitem[Team(2024)]{team2024chameleon}
Chameleon Team.
\newblock Chameleon: Mixed-modal early-fusion foundation models.
\newblock \emph{arXiv preprint arXiv:2405.09818}, 2024.

\bibitem[Tian et~al.(2024)Tian, Jiang, Yuan, Peng, and Wang]{tian2024visual}
Keyu Tian, Yi Jiang, Zehuan Yuan, Bingyue Peng, and Liwei Wang.
\newblock Visual autoregressive modeling: Scalable image generation via next-scale prediction.
\newblock \emph{arXiv preprint arXiv:2404.02905}, 2024.

\bibitem[Touvron et~al.(2023)Touvron, Lavril, Izacard, Martinet, Lachaux, Lacroix, Rozi{\`e}re, Goyal, Hambro, Azhar, et~al.]{touvron2023llama}
Hugo Touvron, Thibaut Lavril, Gautier Izacard, Xavier Martinet, Marie-Anne Lachaux, Timoth{\'e}e Lacroix, Baptiste Rozi{\`e}re, Naman Goyal, Eric Hambro, Faisal Azhar, et~al.
\newblock Llama: Open and efficient foundation language models.
\newblock \emph{arXiv preprint arXiv:2302.13971}, 2023.

\bibitem[Van Den~Oord et~al.(2017)Van Den~Oord, Vinyals, et~al.]{van2017neural}
Aaron Van Den~Oord, Oriol Vinyals, et~al.
\newblock Neural discrete representation learning.
\newblock \emph{Advances in neural information processing systems}, 30, 2017.

\bibitem[Wang et~al.(2004)Wang, Bovik, Sheikh, and Simoncelli]{wang2004image}
Zhou Wang, Alan~C Bovik, Hamid~R Sheikh, and Eero~P Simoncelli.
\newblock Image quality assessment: from error visibility to structural similarity.
\newblock \emph{IEEE transactions on image processing}, 13\penalty0 (4):\penalty0 600--612, 2004.

\bibitem[Wei et~al.(2022)Wei, Fan, Xie, Wu, Yuille, and Feichtenhofer]{wei2022masked}
Chen Wei, Haoqi Fan, Saining Xie, Chao-Yuan Wu, Alan Yuille, and Christoph Feichtenhofer.
\newblock Masked feature prediction for self-supervised visual pre-training.
\newblock In \emph{Proceedings of the IEEE/CVF conference on computer vision and pattern recognition}, pages 14668--14678, 2022.

\bibitem[Xiao et~al.(2024)Xiao, Wang, Zhou, Yuan, Xing, Yan, Wang, Huang, and Liu]{xiao2024omnigen}
Shitao Xiao, Yueze Wang, Junjie Zhou, Huaying Yuan, Xingrun Xing, Ruiran Yan, Shuting Wang, Tiejun Huang, and Zheng Liu.
\newblock Omnigen: Unified image generation.
\newblock \emph{arXiv preprint arXiv:2409.11340}, 2024.

\bibitem[Xie et~al.(2024)Xie, Mao, Bai, Zhang, Wang, Lin, Gu, Chen, Yang, and Shou]{xie2024show}
Jinheng Xie, Weijia Mao, Zechen Bai, David~Junhao Zhang, Weihao Wang, Kevin~Qinghong Lin, Yuchao Gu, Zhijie Chen, Zhenheng Yang, and Mike~Zheng Shou.
\newblock Show-o: One single transformer to unify multimodal understanding and generation.
\newblock \emph{arXiv preprint arXiv:2408.12528}, 2024.

\bibitem[Yan et~al.(2021)Yan, Zhang, Abbeel, and Srinivas]{yan2021videogpt}
Wilson Yan, Yunzhi Zhang, Pieter Abbeel, and Aravind Srinivas.
\newblock Videogpt: Video generation using vq-vae and transformers.
\newblock \emph{arXiv preprint arXiv:2104.10157}, 2021.

\bibitem[Yang et~al.(2024)Yang, Yang, Hui, Zheng, Yu, Zhou, Li, Li, Liu, Huang, et~al.]{yang2024qwen2}
An Yang, Baosong Yang, Binyuan Hui, Bo Zheng, Bowen Yu, Chang Zhou, Chengpeng Li, Chengyuan Li, Dayiheng Liu, Fei Huang, et~al.
\newblock Qwen2 technical report.
\newblock \emph{arXiv preprint arXiv:2407.10671}, 2024.

\bibitem[Yao et~al.(2024)Yao, Yu, Zhang, Wang, Cui, Zhu, Cai, Li, Zhao, He, et~al.]{yao2024minicpm}
Yuan Yao, Tianyu Yu, Ao Zhang, Chongyi Wang, Junbo Cui, Hongji Zhu, Tianchi Cai, Haoyu Li, Weilin Zhao, Zhihui He, et~al.
\newblock Minicpm-v: A gpt-4v level mllm on your phone.
\newblock \emph{arXiv preprint arXiv:2408.01800}, 2024.

\bibitem[Yu et~al.(2021)Yu, Li, Koh, Zhang, Pang, Qin, Ku, Xu, Baldridge, and Wu]{yu2021vector}
Jiahui Yu, Xin Li, Jing~Yu Koh, Han Zhang, Ruoming Pang, James Qin, Alexander Ku, Yuanzhong Xu, Jason Baldridge, and Yonghui Wu.
\newblock Vector-quantized image modeling with improved vqgan.
\newblock \emph{arXiv preprint arXiv:2110.04627}, 2021.

\bibitem[Yu et~al.(2023{\natexlab{a}})Yu, Cheng, Sohn, Lezama, Zhang, Chang, Hauptmann, Yang, Hao, Essa, et~al.]{yu2023magvit}
Lijun Yu, Yong Cheng, Kihyuk Sohn, Jos{\'e} Lezama, Han Zhang, Huiwen Chang, Alexander~G Hauptmann, Ming-Hsuan Yang, Yuan Hao, Irfan Essa, et~al.
\newblock Magvit: Masked generative video transformer.
\newblock In \emph{Proceedings of the IEEE/CVF Conference on Computer Vision and Pattern Recognition}, pages 10459--10469, 2023{\natexlab{a}}.

\bibitem[Yu et~al.(2023{\natexlab{b}})Yu, Lezama, Gundavarapu, Versari, Sohn, Minnen, Cheng, Birodkar, Gupta, Gu, et~al.]{yu2023language}
Lijun Yu, Jos{\'e} Lezama, Nitesh~B Gundavarapu, Luca Versari, Kihyuk Sohn, David Minnen, Yong Cheng, Vighnesh Birodkar, Agrim Gupta, Xiuye Gu, et~al.
\newblock Language model beats diffusion--tokenizer is key to visual generation.
\newblock \emph{arXiv preprint arXiv:2310.05737}, 2023{\natexlab{b}}.

\bibitem[Yu et~al.(2024{\natexlab{a}})Yu, Cheng, Wang, Kumar, Macherey, Huang, Ross, Essa, Bisk, Yang, et~al.]{yu2024spae}
Lijun Yu, Yong Cheng, Zhiruo Wang, Vivek Kumar, Wolfgang Macherey, Yanping Huang, David Ross, Irfan Essa, Yonatan Bisk, Ming-Hsuan Yang, et~al.
\newblock Spae: Semantic pyramid autoencoder for multimodal generation with frozen llms.
\newblock \emph{Advances in Neural Information Processing Systems}, 36, 2024{\natexlab{a}}.

\bibitem[Yu et~al.(2024{\natexlab{b}})Yu, Weber, Deng, Shen, Cremers, and Chen]{yu2024image}
Qihang Yu, Mark Weber, Xueqing Deng, Xiaohui Shen, Daniel Cremers, and Liang-Chieh Chen.
\newblock An image is worth 32 tokens for reconstruction and generation.
\newblock \emph{arXiv preprint arXiv:2406.07550}, 2024{\natexlab{b}}.

\bibitem[Zhang et~al.(2023)Zhang, Li, and Bing]{zhang2023video}
Hang Zhang, Xin Li, and Lidong Bing.
\newblock Video-llama: An instruction-tuned audio-visual language model for video understanding.
\newblock \emph{arXiv preprint arXiv:2306.02858}, 2023.

\bibitem[Zhang et~al.(2018)Zhang, Isola, Efros, Shechtman, and Wang]{zhang2018unreasonable}
Richard Zhang, Phillip Isola, Alexei~A Efros, Eli Shechtman, and Oliver Wang.
\newblock The unreasonable effectiveness of deep features as a perceptual metric.
\newblock In \emph{Proceedings of the IEEE conference on computer vision and pattern recognition}, pages 586--595, 2018.

\bibitem[Zheng et~al.(2022)Zheng, Vuong, Cai, and Phung]{zheng2022movq}
Chuanxia Zheng, Tung-Long Vuong, Jianfei Cai, and Dinh Phung.
\newblock Movq: Modulating quantized vectors for high-fidelity image generation.
\newblock \emph{Advances in Neural Information Processing Systems}, 35:\penalty0 23412--23425, 2022.

\bibitem[Zhou et~al.(2024)Zhou, Yu, Babu, Tirumala, Yasunaga, Shamis, Kahn, Ma, Zettlemoyer, and Levy]{zhou2024transfusion}
Chunting Zhou, Lili Yu, Arun Babu, Kushal Tirumala, Michihiro Yasunaga, Leonid Shamis, Jacob Kahn, Xuezhe Ma, Luke Zettlemoyer, and Omer Levy.
\newblock Transfusion: Predict the next token and diffuse images with one multi-modal model.
\newblock \emph{arXiv preprint arXiv:2408.11039}, 2024.

\bibitem[Zhu et~al.(2023)Zhu, Chen, Shen, Li, and Elhoseiny]{zhu2023minigpt}
Deyao Zhu, Jun Chen, Xiaoqian Shen, Xiang Li, and Mohamed Elhoseiny.
\newblock Minigpt-4: Enhancing vision-language understanding with advanced large language models.
\newblock \emph{arXiv preprint arXiv:2304.10592}, 2023.

\bibitem[Zhu et~al.(2024)Zhu, Wei, and Lu]{zhu2024beyond}
Lei Zhu, Fangyun Wei, and Yanye Lu.
\newblock Beyond text: Frozen large language models in visual signal comprehension.
\newblock In \emph{Proceedings of the IEEE/CVF Conference on Computer Vision and Pattern Recognition}, pages 27047--27057, 2024.

\end{thebibliography}
}

\end{document}